\documentclass[letterpaper]{article} % DO NOT CHANGE THIS
\usepackage{aaai2027}
\usepackage[hyphens]{url}  % DO NOT CHANGE THIS
\usepackage{graphicx} % DO NOT CHANGE THIS
\usepackage{natbib}  % DO NOT CHANGE THIS AND DO NOT ADD ANY OPTIONS TO IT
\usepackage{caption} % DO NOT CHANGE THIS AND DO NOT ADD ANY OPTIONS TO IT
\usepackage{algorithm}
\usepackage{algorithmic}
 
\usepackage{latexsym}
\usepackage{amsmath}
\usepackage{amssymb} % needed for \mathbb, used throughout Sec. Methodology and the appendix
\usepackage{booktabs}
\usepackage{subcaption}
\usepackage[T1]{fontenc}
\usepackage[utf8]{inputenc}
\usepackage{microtype}
\usepackage{inconsolata}
\usepackage{multirow}
\usepackage[table]{xcolor}
\usepackage{tikz}
\usepackage{pgfplots}
\pgfplotsset{compat=1.18}
\usepgfplotslibrary{groupplots}
 
\usepackage{newfloat}
\usepackage{listings}
\DeclareCaptionStyle{ruled}{labelfont=normalfont,labelsep=colon,strut=off} % DO NOT CHANGE THIS
\floatstyle{ruled}
\newfloat{listing}{tb}{lst}{}
\floatname{listing}{Listing}
 
\title{\texttt{Omni2LoRA}: Coherence-Preserving Parametric Memory for Efficient Omni Language Models}
 
\author{
    Puneet Mathur\footnote{Work done at University of Maryland, College Park},
    Manan Suri,
    Dinesh Manocha
}
\affiliations{
    University of Maryland, College Park\\
    \textit{\{puneetm, manans, dmanocha\}@umd.edu}
}

\begin{document}

\maketitle

\begin{abstract}
    Omnimodal language models (OLMs) enable unified audio-visual understanding, but processing long joint token sequences makes inference computationally prohibitive. While recent token compression methods attempt to alleviate this burden, compressing modalities in isolation often destroys the temporal cross-modal anchors necessary for coherent reasoning. We introduce \texttt{\textbf{Omni2LoRA}}, a two-stage framework for efficient parametric memory compression via coherence-preserving context distillation that bypasses the token bottleneck entirely. First, a Perceiver hypernetwork processes intermediate representations from a frozen OLM to encode the multimodal context into a full-rank Low-Rank Adaptation (LoRA) adapter in a single forward pass. To prevent the resulting parameter footprint from scaling linearly with recording length, we optimize a discrete rank allocation policy via Group Relative Policy Optimization (GRPO) that uses a modality-ablated counterfactual reward to explicitly penalize the loss of audio-visual coherence, forcing the model to allocate its fixed sub-linear rank budget to synergistic cross-modal anchors rather than isolated visual features. Across three omnimodal backbones (Qwen2.5-Omni-3B/7B, InteractiveOmni-4B), \texttt{Omni2LoRA} operating at a 30\% rank budget outperforms direct full-context inference and strong token-compression baselines (OmniZip, OMAC, O-MARC) on four audio-visual question answering benchmarks, improving average accuracy by 8-12\% over the strongest baseline and remaining stable under compression ratios as tight as 75\%, where token-pruning methods degrade sharply. By converting multimodal memory into a fixed-budget, reusable parameter state, our method drives answer-time multimodal-token load to zero, cutting per-query Time to First Token (TTFT) by up to 12$\times$ relative to full-context inference and amortizing to under 0.5s after a handful of queries, establishing a robust paradigm for long-context omnimodal memory compression.

\end{abstract}


\section{Introduction}
\label{sec:introduction}

\begin{figure}[t]
    \centering
    \includegraphics[width=0.53\textwidth]{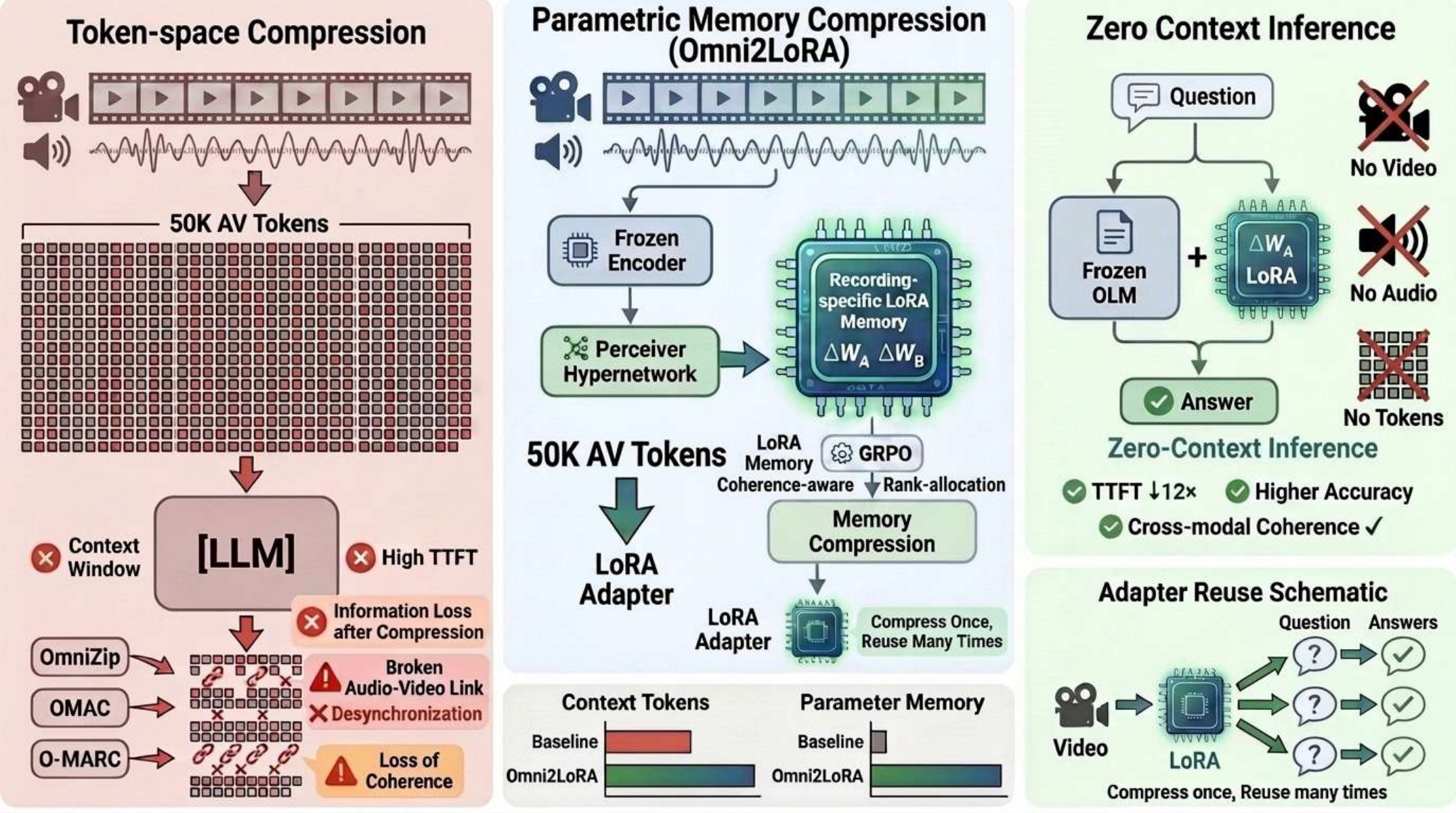}
    \caption{\small{\textbf{Omni2LoRA} internalizes audio--visual recordings into reusable LoRA memory}, enabling zero-context multimodal inference with improved efficiency and cross-modal coherence.}
    \label{fig:teaser}
\end{figure}

Omnimodal large language models have recently emerged as a promising direction for unified audio, visual, and language understanding \cite{cheng2024videollama,xu2025qwen3,liu2026javisgpt}. Unlike vision-language models that primarily rely on visual inputs \cite{wang2024qwen2,liu2023visual}, omnimodal models jointly process video frames, audio streams, and textual instructions, enabling reasoning over both what is seen and what is heard. This capability is particularly important for real-world videos, where speech, ambient sounds, scene transitions, and visual actions often provide complementary evidence for understanding complex events.

Despite these advances, efficient omnimodal inference remains a significant challenge. Omni models encode video frames and their high-frequency audio waveforms into continuous streams of joint visual and acoustic tokens, which are fed directly into the language model's context window. Since audio and video are both token-intensive modalities, their joint representation can easily produce long multimodal sequences, leading to substantial memory overhead and slow inference, especially for long recordings. Pushing models past their capacity threshold often leads to catastrophic degradation, such as losing cross-modal alignment or defaulting to incoherent repetitions \cite{liu2023lost}.

Extensive research has focused on optimizing this context window footprint through token compression or spatial-temporal pruning. Representative approaches include visual token merging \cite{bolya2022token}, attention-guided token pruning \cite{chen2024image}, and audio-guided dynamic compression for omnimodal inputs \cite{tao2025omnizip}. However, extending compression to synchronized audio-visual inputs is non-trivial. Compressing audio and video in isolation can remove cross-modal evidence that is necessary for accurate reasoning \cite{tian2018audio}. A sound becomes meaningful only when grounded in the visible scene, and a visual event may remain ambiguous without its corresponding audio \cite{li2022learning}. Consequently, traditional token compression frameworks often inadvertently sacrifice the very audio-visual coherence they are designed to process, leaving models unable to solve tasks that strictly require joint multimodal evidence.

We introduce a fundamentally distinct approach that bypasses the token context window bottleneck entirely through \textbf{parametric memory compression}. Rather than pruning input tokens to fit within active sequence bounds, we parametrically internalize synchronized multimodal data directly into the model's weight space prior to query generation. Under this paradigm, an input audio-visual recording is mapped into a discrete, instance-specific Low-Rank Adaptation (LoRA) adapter \cite{hu2022lora} generated by a hypernetwork \cite{ha2017hypernetworks} in a single forward pass. Downstream queries are subsequently answered by the frozen model operating solely with these hypernetwork-predicted weights, maintaining \emph{zero audio or visual tokens in the active context window at inference time}.

While parametric internalization successfully bypasses the token-space bottleneck, applying it naively to continuous multimodal streams merely shifts the scalability bottleneck into parameter space. Generating full-rank adapter weights for every temporal chunk across all transformer layers produces an adapter whose size grows linearly with recording length, making parameter storage impractical for long recordings. Furthermore, enforcing a uniform rank allocation across time ignores the highly bursty nature of audio-visual information. As dense visual representations typically exhibit substantially larger activation norms than sparse acoustic cues \cite{wang2020makes}, a naive weight generation process is highly susceptible to \textit{modality collapse}. Under a constrained parameter budget, the hypernetwork disproportionately allocates capacity to visually dominant background features, irreversibly pruning the brief, temporally grounded acoustic anchors required for complex cross-modal reasoning. Consequently, achieving scalable omnimodal memory is not merely a matter of capping the adapter size; it necessitates a dynamic allocation mechanism that is mathematically incentivized to defend audio-visual coherence against visual dominance.

\begin{figure*}[t]
  \centering
  \includegraphics[width=\textwidth]{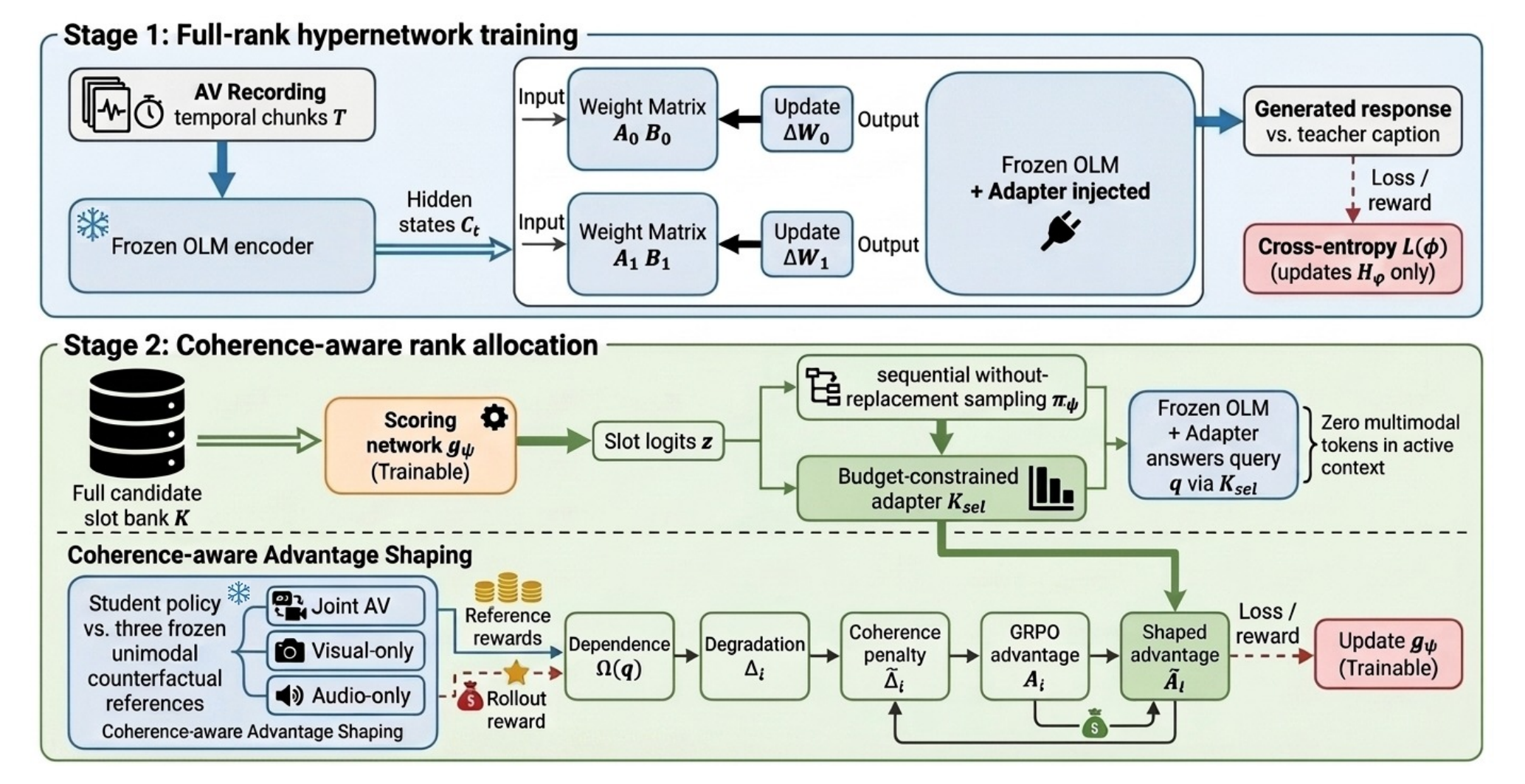}
  \caption{\small{\textbf{Overview of \texttt{Omni2LoRA}.} A synchronized audio--visual recording is first encoded by a frozen omni LM, whose intermediate representations are consumed by a Perceiver hypernetwork to generate a full-rank bank of candidate LoRA slots (\textbf{Stage~1}). A trainable scoring policy then sequentially selects a fixed-budget subset of slots to construct a compact adapter (\textbf{Stage~2}), which is optimized with GRPO using a coherence-aware advantage computed from joint, visual-only, and audio-only frozen reference rollouts, encouraging preservation of cross-modal dependencies under severe compression. The resulting adapter allows the frozen backbone to answer inference queries without retaining multimodal tokens in its context window.}}
  \label{fig:method}
\end{figure*}

\noindent\textbf{Main Result (Fig. \ref{fig:teaser})}: We present \textbf{\texttt{Omni2LoRA}}, a two-stage framework that couples full-rank hypernetwork pretraining with coherence-aware rank allocation to achieve optimal parametric memory compression. First, a hierarchical Perceiver hypernetwork \cite{jaegle2021perceiver} maps synchronized audio-visual streams into layer-wise candidate LoRA weights. Second, we introduce a reinforcement-learned, discrete allocation policy that dynamically determines which subset of the generated rank directions to retain under a fixed, sub-linear adapter budget. To solve the audio-visual coherence problem, this policy is optimized via Group Relative Policy Optimization (GRPO) \cite{shao2024deepseekmath} using a novel advantage shaping mechanism based on unimodal counterfactuals. By penalizing degradation specifically on queries that require the intersection of audio and video, the allocation policy is mathematically coerced into preserving synergistic cross-modal memory. During inference, the multimodal recording is processed exactly once to produce a compact, recording-specific adapter, amortizing the initial encoding cost across all future interactions. Our \textbf{main contributions} are:

\begin{itemize}
    \item \textbf{Parametric AV Memory Compression}: \texttt{Omni2LoRA} is a parametric memory compression framework that internalizes audio-visual context into model weights via a hypernetwork-generated LoRA adapter, enabling omnimodal reasoning with \textit{zero multimodal tokens in context}.
    
    \item \textbf{Coherence-Aware Rank Allocation}: We formulate an RL-driven allocation policy that compresses hypernetwork adapters into a fixed rank budget, using unimodal counterfactuals in the GRPO advantage to preserve mutually dependent audio-visual anchors.
    
    \item \textbf{Strong Performance on Audio-Visual Memory Compression}: \texttt{Omni2LoRA} \textit{outperforms SOTA compression baselines} (OmniZip \cite{tao2025omnizip}, OMAC and O-MARC \cite{wu2026marc}) \textit{by 8-12\% across omnimodal backbones} on UGC-AVQA \cite{wu2026marc}, a benchmark strictly requiring joint acoustic-visual evidence, with consistent gains on WorldSense \cite{hong2025worldsense}, OmniVideoBench \cite{li2025omnivideobench}, and DailyOmni \cite{zhou2025daily}. It maintains high cross-modal coherence across increasing compression ratios and shows stability across frame sampling rates, where token-in-context baselines catastrophically degrade.
    
    \item \textbf{Extreme Inference Efficiency}: Converting multimodal context into a fixed-budget adapter drives answer-time token load to zero, cutting query TTFT by \textit{up to 12$\times$} on VidCapBench \cite{chen2025vidcapbench} and amortizing to \textit{sub-second latency} across repeated queries.
\end{itemize}

\section{Related Work}
\label{sec:related_work}

\paragraph{Omnimodal Token Compression.}
Recent omnimodal large language models (OLMs) jointly process audio and visual streams to enable unified multimodal reasoning, but the resulting token sequences incur substantial computational and memory costs. Existing methods improve inference efficiency through token pruning or merging. Generic token reduction methods, including Token Merging (ToMe)~\cite{bolya2022token} and layer-wise token pruning for vision-language models~\cite{chen2024image}, reduce redundant visual representations but are not designed to preserve cross-modal dependencies. More recent omnimodal approaches explicitly model audio--visual interactions. OmniZip~\cite{tao2025omnizip} performs audio-guided dynamic token compression, while OMAC and its distilled extension O-MARC~\cite{wu2026marc} preserve query-relevant visual memory and temporally aligned acoustic cues during compression. Despite improving efficiency, these methods continue to rely on compressed multimodal tokens in the context window, limiting scalability for long recordings.

\paragraph{Parametric Memory via LoRA.}
An alternative direction replaces token-level storage with parameter-space memory. Recent studies show that Low-Rank Adaptation (LoRA)~\cite{hu2022lora} serves as an effective parametric memory for storing contextual information~\cite{back2026understanding,xu2026lora}, with finite, rank-dependent storage capacity governed by a parametric memory scaling law~\cite{xu2026lora}. To mitigate catastrophic forgetting, recent memory management frameworks organize multiple adapters into modular memory banks with latent routing and retrieval mechanisms~\cite{zheng2026context}. While these approaches demonstrate the feasibility of parameter-space memory, they primarily focus on textual knowledge rather than synchronized audio--visual memory under constrained parameter budgets.

\paragraph{Context-to-LoRA Hypernetworks.}
Hypernetworks~\cite{ha2017hypernetworks} generate task-specific parameters in a single forward pass, providing an efficient alternative to iterative gradient-based adaptation. Recent context-to-LoRA methods directly map contextual inputs into LoRA adapters. Doc2LoRA~\cite{charakorn2026doc} internalizes long documents into parameter space, while Frames2LoRA~\cite{suri2026frames2lora} encodes long videos into LoRA adapters, eliminating visual tokens during inference. We extend this paradigm to synchronized audio--visual reasoning by introducing a GRPO-based rank allocation policy that preserves cross-modal dependencies under a fixed parameter budget.

\section{Methodology}
\label{sec:methodology}

\texttt{Omni2LoRA} (Fig.~\ref{fig:method}) compresses synchronized audio--visual recordings into the parameter space of a frozen OLM through: \textbf{Stage 1: Full-Rank Hypernetwork Training} trains a Perceiver hypernetwork to generate full-rank LoRA adapters from multimodal representations, enabling zero-token inference. While this establishes representational capacity, the generated adapter rank scales linearly with the number of temporal chunks, inadvertently translating the scaling bottleneck into parameter space. To prevent the destruction of cross-modal reasoning, \textbf{Stage 2: Memory-Augmented Compression Distillation} learns a budget-constrained rank allocation policy via reinforcement learning to preserve cross-modal anchors essential for coherent audio--visual reasoning.

\subsection{Problem Formulation}
\label{sec:formulation}

Let $v$ denote a synchronized audio-visual recording, $i$ an internalization instruction, $q$ a downstream text query, and $y$ the target response. Let $E$ denote the frozen OLM encoder and $F$ the frozen OLM answer model. We partition $v$ into $T$ non-overlapping temporal chunks $\{v_1, \dots, v_T\}$. For chunk $v_t$, the encoder jointly processes the multimodal streams to produce layer-wise hidden states:
\begin{equation}
\small
C_t = E(v_t, i) \in \mathbb{R}^{L \times S \times D}
\end{equation}
where $L$ is the number of transformer layers, $S$ is the fused audio-visual sequence length, and $D$ is the hidden dimension. A trainable hypernetwork $H_\phi$ maps $C_t$ to a set of candidate LoRA factors for each target linear module in $F$:
\begin{equation}
\small
\begin{split}
\{A_{\ell,m,t,r}, B_{\ell,m,t,r}\}_{r=1}^{R_{\max}} &= H_\phi(C_t), \\
&\quad \ell \in [1,L],\; m \in [1,M]
\end{split}
\end{equation}
where $\ell$ indexes transformer layers, $m$ indexes target linear modules, and $r$ indexes the $R_{\max}$ rank directions. $H_\phi$ acts as a hierarchical Perceiver-style resampler \cite{jaegle2021perceiver}, utilizing cross-attention to distill the representations into rank factors. For a frozen linear layer $W \in \mathbb{R}^{d_{\text{out}} \times d_{\text{in}}}$, the fully composed adapter across all chunks yields the forward pass:
\begin{equation}
\small
y = xW^\top + \sum_{t=1}^{T} \sum_{r=1}^{R_{\max}} s \, (x A_{\ell,m,t,r}^\top)\, B_{\ell,m,t,r}
\label{eq:lora-inject}
\end{equation}
where $s$ is a fixed scaling factor. During training, $E$ and $F$ remain entirely frozen.

\subsection{Stage 1: Full-Rank Hypernetwork Training}
\label{sec:stage1}

$H_\phi$ is pre-trained via teacher-forced cross-entropy against cached teacher-generated audio-visual captions. For a training instance $(v, i, p, y)$, we instantiate the complete candidate bank of $T \cdot L \cdot M \cdot R_{\max}$ rank-one components and optimize:
\begin{equation}
\small
\mathcal{L}(\phi) = -\sum_{\tau} \log p_\phi\big(y_\tau \mid y_{<\tau}, p, \theta(v)\big)
\label{eq:stage1-loss}
\end{equation}
where $\theta(v)$ denotes the full generated adapter. At convergence, $H_\phi$ is frozen, functioning as a deterministic mapping from an input recording to a comprehensive bank of $|\mathcal{K}| = T \cdot L \cdot M \cdot R_{\max}$ candidate rank-one LoRA updates, called \textit{slots}.

\subsection{Stage 2: Memory-Augmented Compression Distillation}
\label{sec:stage2}

Deploying the full slot bank at inference requires a total rank budget that scales linearly with T, necessitating compression. However, compressing modalities independently risks removing the cross-modal evidence necessary for accurate reasoning \cite{tian2018audio,li2022learning}. To solve this audio-visual coherence problem while respecting a fixed budget, we formulate rank allocation as a compression-distillation task, following the memory-augmented compression framing of prior work \cite{wu2026marc}. We define an allocation $\mathcal{K}_{\text{sel}} \subset \mathcal{K}$ such that $|\mathcal{K}_{\text{sel}}| = R_{\text{total}}$. This hierarchical task requires determining which chunks contain query-relevant information, and which specific rank directions within those chunks efficiently encode cross-modal memory.

\paragraph{Scoring Network.} A lightweight scoring network $g_\psi$ is conditioned on the pooled audio-visual summary of each chunk, $\bar{C}_t = \mathrm{mean}(C_t)$, and a positional embedding of $(\ell, m, r)$ to produce a scalar logit $z_{\ell,m,t,r} = g_\psi(\bar{C}_t, \ell, m, r) \in \mathbb{R}$. $g_\psi$ operates independently of the continuous $A, B$ weights; only $\psi$ is updated in Stage 2.

\paragraph{Sequential Without-Replacement Sampling.} To optimize $g_\psi$ via policy-gradient, allocations are sampled sequentially. At step $k$, conditioned on the selected slots $o_{<k}$, the next slot is drawn via softmax:
\begin{equation}
\pi_\psi(o_k \mid o_{<k}, v) = \frac{\exp(z_{o_k})}{\sum_{j \in \mathcal{K} \setminus o_{<k}} \exp(z_j)}
\label{eq:pl-step}
\end{equation}
The probability of the full allocation is $\pi_\psi(\mathcal{K}_{\text{sel}} \mid v) = \prod_{k=1}^{R_{\text{total}}} \pi_\psi(o_k \mid o_{<k}, v)$. This autoregressive factorization ensures clipped importance-ratio training remains mathematically well-defined for parameter sub-selection.

\noindent\textbf{Reward Function.} Each rollout is scored using a combination of accuracy and format adherence: $R = \frac{1}{2}\, R_{\text{acc}} + \frac{1}{2}\, R_{\text{fmt}}$, both computed via lightweight rule-based checks rather than an auxiliary judge model, keeping per-rollout reward computation cheap during RL sampling.

\subsubsection{Coherence-Aware Advantage Shaping}
\label{sec:shaping}

Optimizing $g_\psi$ solely to maximize the base reward $R$ risks an allocation collapse toward visually dominant features, as visual tokens typically dominate intermediate representation norms. This can inadvertently prune subtle acoustic rank directions, destroying the cross-modal evidence necessary for accurate reasoning. To explicitly penalize the loss of audio-visual coherence, we introduce a unimodal counterfactual advantage shaping mechanism. For every training query $q$ over recording $v$, we evaluate the frozen answer model $F$ (without the adapter) under three distinct full-token context baselines to precompute fixed reference rewards, cached once and reused throughout Stage 2 training since $F$ remains frozen: \textbf{(i) Joint AV Reference ($R^{\text{AV}}$):} $F$ answers $q$ utilizing the full synchronized audio-visual context. \textbf{(ii) Visual-Only Reference ($R^{\text{V}}$):} $F$ answers $q$ with the audio stream masked. \textbf{(iii) Audio-Only Reference ($R^{\text{A}}$):} $F$ answers $q$ with the video frames masked.

\noindent We quantify the intrinsic cross-modal necessity of query $q$ via the \textit{Audio-Visual Dependence Score} $\Omega(q)$, defined as the performance degradation under optimal unimodal context:
\begin{equation}
\small
\Omega(q) = \mathrm{ReLU} \Big( R^{\text{AV}} - \max(R^{\text{V}}, R^{\text{A}}) \Big)
\end{equation}
A high $\Omega(q)$ isolates queries that strictly require the synthesis of both modalities. Next, we evaluate the allocated student branch. $G$ allocations are sampled from $\pi_\psi(\cdot \mid v)$. For each, $F$ answers $q$ using strictly the budget-constrained adapter, yielding a rollout reward $R_i^{\text{alloc}}$. We calculate the base parametric degradation score $\Delta_i$, where $\tau$ is small constant for numerical stability:
\begin{equation}
\small
\Delta_i = \frac{\mathrm{ReLU}\big(R^{\text{AV}} - R_i^{\text{alloc}}\big)}{|R^{\text{AV}}| + \tau}
\end{equation}
To enforce coherence, we construct a penalty multiplier that amplifies the degradation gap for highly cross-modal queries:
\begin{equation}
\small
\tilde{\Delta}_i = \Delta_i \cdot \big(1 + \eta \, \Omega(q)\big)
\end{equation}
where $\eta$ governs the coherence penalty strength. Let $\hat{A}_i$ denote the standard group-relative GRPO advantage \cite{shao2024deepseekmath} of rollout $i$. We reshape this advantage:
\begin{equation}
\small
w_i = \mathrm{ReLU}(\hat{A}_i) \cdot \tilde{\Delta}_i, \qquad \tilde{A}_i = \hat{A}_i - \lambda w_i
\label{eq:shaped-advantage}
\end{equation}
Allocations that score well by chance but drop critical cross-modal rank directions on highly dependent queries receive a severely penalized update, forcing the policy to preserve audio-visual memory.

\noindent\textbf{Policy Optimization.} We optimize $\pi_\psi$ employing a clipped, KL-regularized policy-gradient objective in the style of PPO \cite{schulman2017proximal}:

\begin{equation}
\small
\mathcal{L}_{\text{O2L-GRPO}} = - \mathbb{E}\left[ \frac{1}{G} \sum_{i=1}^{G} \rho_i(\psi)\, \tilde{A}_i \; - \; \beta\, \mathrm{KL}\big(\pi_\psi \,\|\, \pi_{\psi_{\text{ref}}}\big) \right]
\label{eq:final-obj}
\end{equation}
where $\rho_i(\psi)$ is the clipped policy ratio and $\pi_{\psi_{\text{ref}}}$ is the reference allocation policy. $H_\phi$, $E$, and $F$ remain frozen.

\subsubsection{Inference}

During inference, \texttt{Omni2LoRA} processes the multimodal recording exactly once. The frozen $H_\phi$ constructs the full slot bank $\mathcal{K}$; the frozen $g_\psi$ scores every slot and greedily selects the top-$R_{\text{total}}$ to fix the allocation $\mathcal{K}_{\text{sel}}$. This budget-constrained adapter is reused for all subsequent queries. $F$ answers with zero multimodal tokens in its active context window, converting encoding overhead into an amortized one-time setup cost.

\begin{table*}[t]
\centering
\resizebox{\textwidth}{!}{
\begin{tabular}{l c c c c c c c c c c c}
\toprule
\multirow{2}{*}{Method} & \multicolumn{2}{c}{Settings} & \multicolumn{1}{c}{DailyOmni} & \multicolumn{5}{c}{UGC-AVQA} & \multicolumn{1}{c}{OmniVideo} & \multicolumn{1}{c}{WorldSense} & \multicolumn{1}{c}{Average} \\
\cmidrule(lr){2-3} \cmidrule(lr){4-4} \cmidrule(lr){5-9} \cmidrule(lr){10-10} \cmidrule(lr){11-11} \cmidrule(lr){12-12}
& Retained Ratio & Frames & Overall Accuracy & AVEP & AVST & CSAVA & FGAVC & Overall Accuracy (\%) & Avg. Score & Overall Accuracy & Acc (\%) \\
\midrule
\multicolumn{12}{c}{\textit{Qwen2.5-Omni-3B}} \\
\midrule
Full Tokens & 100\% & 32 & 53.8 & 44.7 & 52.4 & 48.3 & 50.5 & 48.9 & 31.6 & 42.2 & 44.1 \\
OmniZip & 30\% & 32 & 47.7 & 44.7 & 49.0 & 44.9 & 49.8 & 47.1 & 29.3 & 39.7 & 41.0 \\
OMAC & 30\% & 32 & 49.9 & 47.6 & 52.4 & 47.1 & 49.5 & 49.1 & 30.8 & 41.2 & 42.8 \\
O-MARC & 30\% & 32 & 52.3 & 52.4 & 61.4 & 55.8 & 61.9 & 57.9 & 31.0 & 42.1 & 45.8 \\
\rowcolor{cyan!20}\texttt{Omni2LoRA} & 30\% & 32 & \textbf{53.6}$^{\dagger}$ & \textbf{54.2} & \textbf{64.3} & \textbf{60.5} & \textbf{65.7} & \textbf{61.7}$^{\dagger}$ & \textbf{32.5}$^{\dagger}$ & \textbf{44.0}$^{\dagger}$ & \textbf{47.3}$^{\dagger}$\\
\midrule
\multicolumn{12}{c}{\textit{InteractiveOmni-4B}} \\
\midrule
Full Tokens & 100\% & 32 & 55.1 & 51.3 & 52.1 & 49.8 & 56.8 & 52.5 & 34.2 & 42.2 & 46.0 \\
OmniZip & 30\% & 32 & 49.9 & 50.3 & 50.6 & 49.5 & 53.9 & 51.0 & 30.0 & 39.6 & 42.6 \\
OMAC & 30\% & 32 & 52.1 & 51.2 & 54.9 & 49.1 & 55.7 & 52.7 & 30.9 & 41.0 & 44.1 \\
O-MARC & 30\% & 32 & 54.5 & 54.9 & 58.2 & 51.3 & 57.3 & 55.4 & 32.2 & 41.2 & 45.8 \\
\rowcolor{cyan!20}\texttt{Omni2LoRA} & 30\% & 32 & \textbf{56.0}$^{\dagger}$ & \textbf{57.8} & \textbf{62.5} & \textbf{54.3} & \textbf{61.8} & \textbf{59.1}$^{\dagger}$ & \textbf{33.6}$^{\dagger}$ & \textbf{43.3}$^{\dagger}$ & \textbf{47.6}$^{\dagger}$ \\
\midrule
\multicolumn{12}{c}{\textit{Qwen2.5-Omni-7B}} \\
\midrule
Full Tokens & 100\% & 32 & 56.3 & 52.7 & 54.4 & 51.2 & 58.3 & 54.1 & 34.6 & 43.6 & 47.2 \\
OmniZip & 30\% & 32 & 51.8 & 51.5 & 53.6 & 50.0 & 55.8 & 52.7 & 30.0 & 39.6 & 43.5 \\
OMAC & 30\% & 32 & 53.6 & 52.4 & 54.9 & 49.3 & 56.3 & 53.2 & 30.9 & 42.4 & 45.0 \\
O-MARC & 30\% & 32 & 60.4 & 65.7 & 68.2 & 58.3 & 66.3 & 64.6 & 35.2 & 44.0 & 51.1 \\
\rowcolor{cyan!20}\texttt{Omni2LoRA} & 30\% & 32 & \textbf{63.6}$^{\dagger}$ & \textbf{68.2} & \textbf{70.4} & \textbf{63.5} & \textbf{70.2} & \textbf{68.0}$^{\dagger}$ & \textbf{36.6}$^{\dagger}$ & \textbf{45.8}$^{\dagger}$ & \textbf{53.2}$^{\dagger}$ \\
\midrule
\bottomrule
\end{tabular}
}
\caption{\small{\textbf{Comparison with state-of-the-art token compression baselines.} Under the same frame budget, \texttt{Omni2LoRA} consistently outperforms full-token inference, OmniZip, OMAC, and O-MARC across three omnimodal backbones. DailyOmni and WorldSense report accuracy, OmniVideo reports average score, and UGC-AVQA reports category-wise and overall accuracy. Statistical significance compared to the best-performing baseline (O-MARC) is denoted by $^{\dagger}$ for $p < 0.05$ under Wilcoxon Signed Rank test.}}
\label{tab:main_comparison}
\end{table*}

\begin{table*}[t]
\centering
\small
\resizebox{\textwidth}{!}{
\begin{tabular}{l c c c c c c c c c c c}
\toprule
\multirow{2}{*}{Method} & \multicolumn{2}{c}{Settings} & \multicolumn{1}{c}{DailyOmni} & \multicolumn{5}{c}{UGC-AVQA} & \multicolumn{1}{c}{OmniVideo} & \multicolumn{1}{c}{WorldSense} & \multicolumn{1}{c}{Average} \\
\cmidrule(lr){2-3} \cmidrule(lr){4-4} \cmidrule(lr){5-9} \cmidrule(lr){10-10} \cmidrule(lr){11-11} \cmidrule(lr){12-12}
& Retained Ratio & Frames & Overall Accuracy & AVEP & AVST & CSAVA & FGAVC & Overall Accuracy & Avg. Score & Overall Accuracy & ACC \\
\midrule
\multicolumn{12}{c}{\textit{Qwen2.5-Omni-3B}} \\
\midrule
Direct AV-in-context & 100\% & 32 & 53.8 & 44.7 & 52.4 & 48.3 & 50.5 & 48.9 & 31.6 & 42.2 & 44.1 \\
Full-rank Adapter & 100\% & 32 & 52.5 & 46.1 & 54.0 & 48.9 & 52.1 & 50.3 & 31.0 & 41.5 & 43.8 \\
\midrule
\rowcolor{red!40}Uniform Allocation & 30\% & 32 & 45.2 $\pm$ 1.4 & 41.5 & 46.8 & 43.2 & 48.1 & 44.9 $\pm$ 1.6 & 28.1 $\pm$ 1.2 & 38.6 $\pm$ 1.1 & 39.2 $\pm$ 1.3 \\
\rowcolor{red!20}Norm-scored Allocation & 30\% & 32 & 48.2 $\pm$ 1.1 & 45.0 & 50.1 & 46.5 & 50.0 & 47.9 $\pm$ 1.2 & 29.8 $\pm$ 0.9 & 40.1 $\pm$ 0.8 & 41.5 $\pm$ 1.0 \\
\textbf{O2L-GRPO (Ours)} & 30\% & 32 & \textbf{53.6} $\pm$ 0.4 & \textbf{54.2} & \textbf{64.3} & \textbf{60.5} & \textbf{65.7} & \textbf{61.7} $\pm$ 0.5 & \textbf{31.5} $\pm$ 0.3 & \textbf{42.2} $\pm$ 0.4 & \textbf{47.3} $\pm$ 0.4\\
\midrule
\multicolumn{12}{c}{\textit{InteractiveOmni-4B}} \\
\midrule
Direct AV-in-context & 100\% & 32 & 55.1 & 51.3 & 52.1 & 49.8 & 56.8 & 52.5 & 34.2 & 42.2 & 46.0 \\
Full-rank Adapter & 100\% & 32 & 54.3 & 50.5 & 52.5 & 49.2 & 56.0 & 52.1 & 33.5 & 41.8 & 45.4 \\
\midrule
\rowcolor{red!40}Uniform Allocation & 30\% & 32 & 46.8 $\pm$ 1.5 & 47.1 & 47.5 & 45.5 & 50.2 & 47.6 $\pm$ 1.4 & 29.6 $\pm$ 1.1 & 39.0 $\pm$ 1.2 & 40.7 $\pm$ 1.3 \\
\rowcolor{red!20}Norm-scored Allocation & 30\% & 32 & 50.4 $\pm$ 1.0 & 49.8 & 50.2 & 48.0 & 53.1 & 50.3 $\pm$ 1.1 & 31.2 $\pm$ 0.9 & 40.5 $\pm$ 0.9 & 43.1 $\pm$ 1.0 \\
\textbf{O2L-GRPO (Ours)} & 30\% & 32 & \textbf{55.0} $\pm$ 0.4 & \textbf{57.8} & \textbf{62.5} & \textbf{54.3} & \textbf{61.8} & \textbf{59.1} $\pm$ 0.5 & \textbf{33.6} $\pm$ 0.4 & \textbf{42.8} $\pm$ 0.4 & \textbf{47.6} $\pm$ 0.4 \\
\midrule
\multicolumn{12}{c}{\textit{Qwen2.5-Omni-7B}} \\
\midrule
Direct AV-in-context & 100\% & 32 & 56.3 & 52.7 & 54.4 & 51.2 & 58.3 & 54.1 & 34.6 & 43.6 & 47.2 \\
Full-rank Adapter & 100\% & 32 & 55.8 & 53.5 & 55.1 & 52.0 & 58.0 & 54.7 & 34.2 & 43.1 & 47.0 \\
\midrule
\rowcolor{red!40}Uniform Allocation & 30\% & 32 & 48.5 $\pm$ 1.6 & 49.1 & 50.4 & 47.8 & 52.5 & 49.9 $\pm$ 1.5 & 30.2 $\pm$ 1.2 & 40.0 $\pm$ 1.3 & 42.1 $\pm$ 1.4 \\
\rowcolor{red!20}Norm-scored Allocation & 30\% & 32 & 52.5 $\pm$ 1.2 & 51.0 & 52.8 & 49.5 & 55.1 & 52.1 $\pm$ 1.1 & 31.8 $\pm$ 0.9 & 41.2 $\pm$ 1.0 & 44.4 $\pm$ 1.1 \\
\textbf{O2L-GRPO (Ours)} & 30\% & 32 & \textbf{63.6} $\pm$ 0.3 & \textbf{68.2} & \textbf{70.4} & \textbf{63.5} & \textbf{70.2} & \textbf{68.0} $\pm$ 0.4 & \textbf{36.6} $\pm$ 0.3 & \textbf{44.8} $\pm$ 0.4 & \textbf{53.2} $\pm$ 0.3 \\
\bottomrule
\end{tabular}
}
\caption{\small{\textbf{Ablation analysis} of \texttt{Omni2LoRA} establishes non-inferiority to the \textit{Direct AV-in-context} baseline, and demonstrates the representational capacity of the uncompressed \textit{Full-rank Adapter}. We compare our \textit{O2L-GRPO} allocation policy against \textit{Uniform Allocation} and magnitude-based \textit{Norm-scored Allocation} under a strict matched compression budget. We report 95\% empirical bootstrap CI for statistical stability.}}
\label{tab:main_results}
\end{table*}

%%%%%%%%%%%

\begin{figure*}[t]
\centering
\small
% --- Invisible plot that generates the shared legend ---
\begin{tikzpicture}
\begin{axis}[
    hide axis,
    xmin=0, xmax=1, ymin=0, ymax=1,
    legend columns=5,
    legend style={
        draw=none,
        font=\footnotesize,
        column sep=0.3cm,
        /tikz/every even column/.append style={column sep=0.3cm}
    },
    legend to name=sharedlegend,
]
\addplot[color=gray, dashed, thick] coordinates {(0,0)};
\addlegendentry{Direct AV-in-context}
\addplot[color=blue, mark=square*, thick] coordinates {(0,0)};
\addlegendentry{OmniZip}
\addplot[color=orange, mark=triangle*, thick] coordinates {(0,0)};
\addlegendentry{OMAC}
\addplot[color=red, mark=diamond*, thick] coordinates {(0,0)};
\addlegendentry{OMARC}
\addplot[color=green, mark=*, thick] coordinates {(0,0)};
\addlegendentry{Ours}
\end{axis}
\end{tikzpicture}

\ref{sharedlegend}
% \vspace{0.2cm}

\begin{subfigure}[b]{0.48\textwidth}
    \centering
    \begin{tikzpicture}
    \begin{axis}[
        width=\linewidth,
        height=3.5cm,
        xlabel={Compression Ratio (\%)},
        ylabel={Overall Accuracy},
        xmin=20, xmax=80,
        ymin=45, ymax=65,
        xtick={25, 50, 75},
        ytick={45, 50, 55, 60, 65},
        grid=major,
        grid style={dashed, gray!30},
        thick,
        tick label style={font=\footnotesize},
        label style={font=\footnotesize}
    ]
    \addplot[color=gray, dashed, thick] coordinates {(20, 48.9) (80, 48.9)};
    \addplot[color=blue, mark=square*, thick] coordinates {
        (25, 48.7) (50, 47.8) (75, 47.1)
    };
    \addplot[color=orange, mark=triangle*, thick] coordinates {
        (25, 49.8) (50, 49.5) (75, 49.1)
    };
    \addplot[color=red, mark=diamond*, thick] coordinates {
        (25, 58.1) (50, 57.6) (75, 56.3)
    };
    \addplot[color=green, mark=*, thick] coordinates {
        (25, 62.1) (50, 61.6) (75, 60.7)
    };
    \end{axis}
    \end{tikzpicture}
    \caption{\small{Effect of compression ratio on UGC-AVQA overall accuracy. \texttt{Omni2LoRA} maintains high cross-modal coherence across increasing compression ratios, outperforming token pruning baselines like OMAC, OMARC and OmniZip under strict memory constraints.}}
    \label{fig:comp_ratio}
\end{subfigure}
\hfill
\begin{subfigure}[b]{0.48\textwidth}
    \centering
    \begin{tikzpicture}
    \begin{axis}[
        width=\linewidth,
        height=3.5cm,
        xmode=log,
        log basis x=2,
        log ticks with fixed point,
        xlabel={Number of Frames},
        ylabel={Average Score},
        xmin=5, xmax=1030,
        ymin=20, ymax=50,
        xtick={8, 32, 64, 128, 512, 1024},
        ytick={20, 30, 40, 50},
        grid=major,
        grid style={dashed, gray!30},
        thick,
        tick label style={font=\footnotesize},
        label style={font=\footnotesize}
    ]
    \addplot[smooth, color=gray, dashed, thick, mark=*] coordinates {
        (8, 40.5) (32, 42.2) (64, 41.6) (128, 40.8) (512, 32.5) (1024, 22.0)
    };
    \addplot[smooth, color=blue, mark=square*, thick] coordinates {
        (8, 38.5) (32, 39.7) (64, 40.2) (128, 38.2) (512, 35.0) (1024, 28.5)
    };
    \addplot[smooth, color=orange, mark=triangle*, thick] coordinates {
        (8, 39.8) (32, 41.2) (64, 41.7) (128, 42.5) (512, 38.2) (1024, 32.0)
    };
    \addplot[smooth, color=red, mark=diamond*, thick] coordinates {
        (8, 41.0) (32, 42.1) (64, 42.7) (128, 43.2) (512, 40.5) (1024, 35.5)
    };
    \addplot[smooth, color=green, mark=*, thick] coordinates {
        (8, 41.2) (32, 42.2) (64, 43.1) (128, 44.5) (512, 45.8) (1024, 46.2)
    };
    \end{axis}
    \end{tikzpicture}
    \caption{\small{Effect of video scaling (8 to 1024 frames). Direct AV in-context baseline suffers catastrophic degradation due to context window exhaustion. \texttt{Omni2LoRA} outperforms traditional token compression methods via parametric internalization of audio-visual sequences.}}
    \label{fig:frame_ablation}
\end{subfigure}
\end{figure*}

%%%%%%%%%%%%%
\begin{figure*}[t]
\centering
 
% ---------------------------------------------------
% Shared legend (built from panel (a)'s entries, invoked here)
% NOTE: requires two LaTeX compiles to resolve, like any \ref target.
% ---------------------------------------------------
\ref{fig:common-legend}
\vspace{0.1cm}
 
% ---------------------------------------------------
% (a) Bar Chart: Single-question average TTFT
% ---------------------------------------------------
\scalebox{0.9}{
\begin{minipage}{0.38\textwidth}
\centering
\begin{tikzpicture}
\begin{axis}[
    width=6cm,
    height=4cm,
    ybar,
    bar width=7pt,
    enlarge x limits=0.7,
    ylabel={TTFT (s/question)},
    ymin=0, ymax=6.5,
    ytick={0,2,4,6},
    symbolic x coords={3B, 7B},
    xtick=data,
    xticklabel style={font=\small, align=center},
    tick label style={font=\small},
    ylabel style={font=\small},
    nodes near coords,
    nodes near coords style={font=\tiny, rotate=90, anchor=west},
    nodes near coords align={vertical},
    axis x line*=bottom,
    axis y line*=left,
    ymajorgrids=true,
    grid style={dashed, gray!30},
    clip=false,
    % --- this axis is the single source of truth for the shared legend ---
    legend columns=-1,
    legend to name=fig:common-legend,
    legend style={draw=none, fill=none, font=\small},
]
 
\addplot[fill=gray!70!black, draw=none] coordinates {
    (3B, 5.42)
    (7B, 6.03)
};
\addlegendentry{Base}
 
\addplot[fill=blue!70!black, draw=none] coordinates {
    (3B, 3.83)
    (7B, 4.30)
};
\addlegendentry{OmniZip}
 
\addplot[fill=orange!70!black, draw=none] coordinates {
    (3B, 3.85)
    (7B, 4.32)
};
\addlegendentry{OMAC}
 
\addplot[fill=red!70!black, draw=none] coordinates {
    (3B, 3.19)
    (7B, 3.45)
};
\addlegendentry{OMARC}
 
\addplot[fill=green!80!red, draw=none] coordinates {
    (3B, 0.43)
    (7B, 0.49)
};
\addlegendentry{Omni2LoRA}
 
\end{axis}
\end{tikzpicture}
\caption*{\small{(a) Single-question average TTFT with time taken to internalize the video accounted.}}
\end{minipage}%
}
\hfill
% ---------------------------------------------------
% (b) Line Charts: Amortized TTFT (Groupplot), same 5 methods, same colors
% ---------------------------------------------------
\begin{minipage}{0.58\textwidth}
\centering
\begin{tikzpicture}
\begin{groupplot}[
    group style={
        group size=2 by 1,
        horizontal sep=1.2cm,
        y descriptions at=edge left
    },
    width=0.52\textwidth,
    height=4.2cm,
    xlabel={Questions per video},
    xmin=0, xmax=26,
    ymin=-0.5, ymax=7.5,
    xtick={5,10,15,20,25},
    ytick={0,2,4,6,8},
    axis x line*=bottom,
    axis y line*=left,
    ymajorgrids=true,
    grid style={dashed, gray!30},
]
 
% ================= Plot 1: 3B Model =================
\nextgroupplot[ylabel={Mean TTFT (s/q)}]
 
% --- Base: shaded CI band + line ---
\addplot[color=gray!70!black, very thick, forget plot] coordinates {
    (0,5.49) (5,5.43) (10,5.28) (15,5.89) (20,6.0) (25,6.24)
};
 
% --- OmniZip: line only ---
\addplot[color=blue!70!black, very thick, forget plot] coordinates {
    (0,4.20) (2,4.40) (5,4.60) (10,5.0) (15,5.1) (20,5.15) (25,5.3)
};
 
% --- OMAC: line only ---
\addplot[color=orange!70!black, very thick, forget plot] coordinates {
    (0,3.96) (2,3.72) (5,3.64) (10,4.06) (15,4.16) (20,4.21) (25,4.33)
};
 
% --- OMARC: line only ---
\addplot[color=red!70!black, very thick, forget plot] coordinates {
    (0,3.56) (2,3.41) (5,3.35) (10,3.84) (15,4.00) (20,3.99) (25,4.13)
};
 
% --- O2L: shaded CI band + line ---
\addplot[color=green!80!red, very thick, forget plot] coordinates {
    (0,1.16) (2,0.95) (5,0.72) (10,0.50) (15,0.36) (20,0.36) (25,0.43)
};
 
\node[anchor=north west, font=\bfseries] at (axis cs: 1, 6.7) {3B};
 
% ================= Plot 2: 7B Model =================
\nextgroupplot
 
% --- Base: shaded CI band + line ---
\addplot[color=gray!70!black, very thick, forget plot] coordinates {
    (0,6.29) (5,6.33) (10,6.28) (15,6.69) (20,7.1) (25,7.24)
};
 
% --- OmniZip: line only ---
\addplot[color=blue!70!black, very thick, forget plot] coordinates {
    (0,4.30) (2,4.50) (5,4.550) (10,5.3) (15,5.4) (20,5.55) (25,5.7)
};

% --- OMAC: line only ---
\addplot[color=orange!70!black, very thick, forget plot] coordinates {
    (0,3.99) (2,3.82) (5,3.74) (10,4.16) (15,4.26) (20,4.31) (25,4.43)
};
 
% --- OMARC: line only ---
\addplot[color=red!70!black, very thick, forget plot] coordinates {
    (0,3.66) (2,3.51) (5,3.45) (10,3.94) (15,4.10) (20,4.09) (25,4.23)
};

% --- O2L: shaded CI band + line ---
\addplot[color=green!80!red, very thick, forget plot] coordinates {
    (0,1.36) (2,1.05) (5,0.82) (10,0.70) (15,0.56) (20,0.56) (25,0.43)
};
 
\node[anchor=north west, font=\bfseries] at (axis cs: 1, 7.7) {7B};
 
\end{groupplot}
\end{tikzpicture}
\vspace{0.2cm}
\caption*{\small{(b) Amortized TTFT per question vs. \# of questions per video.}}
\end{minipage}
 
\caption{\small{Inference efficiency of \texttt{Omni2LoRA} compared with base models (Qwen-2.5 3B and 7B variants) across compression baselines (OmniZip, OMAC, OMARC) on VidCapBench.}}
\label{fig:amortization}
\end{figure*}

\section{Experimental Setup}
\label{sec:experiments}

\paragraph{\underline{Models and Training}.} We evaluate \texttt{Omni2LoRA} on three omnimodal backbones: \textbf{Qwen2.5-Omni-3B}, \textbf{InteractiveOmni-4B}, and \textbf{Qwen2.5-Omni-7B}. The audio-visual encoder $E$ and answer model $F$ are initialized from frozen checkpoints. Stage 1  (Full-Rank Hypernetwork Training) utilizes VALOR-1M \cite{liu2024valor} for teacher-forced training, supervised by offline captions. Stage 2 (O2L-GRPO) executes memory-augmented compression distillation on the FineVideo \cite{farre2024finevideo} corpus by drawing queries and video references from downstream training splits to align the policy. All hyperparameters, including $R_{\text{total}}$, are tuned on a held-out validation split.

\paragraph{\underline{Evaluation Benchmarks}.} To assess audio-visual memory retention, cross-modal coherence, and inference efficiency, we evaluate our method on five diverse benchmarks:

\begin{itemize}
    \item \textbf{UGC-AVQA} \cite{wu2026marc}: Audio-visual Question Answering benchmark requiring both acoustic and visual evidence. It targets audio-visual association and evaluates categories like Event Progression, Scene or Temporal Transition, Cross-Scene Audio-Visual Alignment, and Fine-Grained Audio-Visual Contrast.
    \item \textbf{WorldSense} \cite{hong2025worldsense}: Evaluates broad, real-world omnimodal understanding.
    \item \textbf{OmniVideoBench} \cite{li2025omnivideobench}: Emphasizes and evaluates long video reasoning.
    \item \textbf{DailyOmni} \cite{zhou2025daily}: Evaluates video understanding on life scenarios with cross-modal information.
    \item \textbf{VidCapBench} \cite{chen2025vidcapbench}: Provides a rigorous setting to evaluate inference efficiency by associating multiple sequential queries with a single video context.
\end{itemize}

\paragraph{\underline{Baselines}.} We compare our proposed memory-augmented compression distillation (\textit{O2L-GRPO}) against direct inference and three prominent token compression baselines: \textbf{(i) Direct audio-visual-in-context (Full Tokens)}: The frozen backbone utilizing full multimodal context without an adapter. \textbf{(ii) OmniZip} \cite{tao2025omnizip}:  An audio-guided dynamic token compression baseline. \textbf{(iii) OMAC} \cite{wu2026marc}: A training-free plug-in compression method that utilizes query-relevant visual frames and intra-frame contrast tokens to temporally ground acoustic tokens with visual memory. \textbf{(iv) O-MARC} \cite{wu2026marc}: A compression distillation framework that explicitly penalizes compression-induced information loss.

\paragraph{\underline{Ablation settings}.} We compare our proposed coherence-aware rank allocation policy against: \textbf{(i) Full-rank Adapter:} The uncompressed candidate bank retaining all adapter ranks. \textbf{(ii) Uniform Allocation:} Blind allocation policy that retains a fixed budget of slots uniformly without coherence awareness. \textbf{(iii) Norm-scored Allocation:} Heuristic baseline retaining the top-$R_{\text{total}}$ slots based on Frobenius norm of $\lVert A \rVert_F \cdot \lVert B \rVert_F$.

\paragraph{\underline{Metrics}.}  We evaluate output quality using answer accuracy (\%) and average score, depending on the specific benchmark's standard metric. All baselines as well as \texttt{Omni2LoRA} utilize identical videos, prompts, and decoding configurations. To quantify efficiency and robustness, we evaluate performance by varying compression ratios (scaling from 25\% to 75\%), video frames (8 to 1024), and inference-time query Time to First Token (TTFT). More details in Appendix.

\section{Results}
\label{sec:results}

\paragraph{Main Results.} We evaluate \texttt{Omni2LoRA} against leading multimodal token compression frameworks to determine if parametric memory compression offers a superior efficiency-performance trade-off. Table~\ref{tab:main_comparison} shows that internalizing multimodal context directly into parameter space consistently outperforms in-context token reduction under strict retention budgets. Across all three backbone architectures, \texttt{Omni2LoRA} surpasses both contemporary compression methods (OMAC, O-MARC) and the uncompressed \textit{Full Tokens} baseline. For instance, on Qwen2.5-Omni-7B, our approach achieves a 53.2 average accuracy, exceeding O-MARC (51.1) and the full-context ceiling (47.2). These improvements are most pronounced on the UGC-AVQA benchmark, where preserving fine-grained cross-modal coherence is strictly required. By achieving 68.0\% overall accuracy on this dataset, \texttt{Omni2LoRA} proves that parametric memory compression effectively bypasses the cross-modal information loss inherent to spatial-temporal token pruning.

\paragraph{Ablation Analysis.} Table~\ref{tab:main_results} isolates the contribution of our coherence-aware rank allocation against uncompressed ceilings and heuristic baselines. Operating at a 100\% parameter budget, the \textit{Full-rank adapter} achieves near parity with \textit{Direct AV-in-context} inference, validating the hypernetwork's capacity to encode multimodal memory.However, enforcing a naive 30\% budget constraint exposes the fragility of undirected compression: blind \textit{Uniform Allocation} triggers severe degradation across all downstream tasks, while magnitude-based \textit{Norm-scored Allocation} fails to protect subtle cross-modal anchors, struggling heavily on the strict UGC-AVQA benchmark. In stark contrast, the \textit{O2L-GRPO} policy entirely bridges this gap. By learning to allocate ranks specifically where audio and video intersect, it filters out query-irrelevant noise, allowing a sparse 30\% subset to consistently outperform dense uncompressed representations. This confirms that RL-driven advantage shaping is essential to defend audio-visual coherence against visual dominance under tight memory constraints.

\paragraph{Compression Ratio.} We evaluate representational robustness under increasingly constrained memory budgets by scaling compression ratios (25\%, 50\%, 75\%) on UGC-AVQA (Figure \ref{fig:comp_ratio}). As the available capacity decreases, traditional spatial-temporal token pruning techniques struggle to maintain the necessary cross-modal evidence. At extreme 75\% compression, OmniZip degrades to 47.1, direct full-context baseline (48.9). While OMAC and OMARC exhibit better resilience, they still suffer noticeable drops to 49.1 and 56.3, respectively. In stark contrast, \texttt{Omni2LoRA} demonstrates exceptional stability, maintaining accuracy of 60.7 even at 75\% compression. This confirms that our explicitly learned allocation policy successfully prevents independent modality collapse, proving parametric internalization is vastly superior for preserving task-relevant, cross-modal anchors under severe budget constraints.

\paragraph{Video Frames Scaling.} We investigate the representational robustness under temporal scaling by varying the number of sampled video frames from 8 to 1024. Figure \ref{fig:frame_ablation} shows that 
increasing the visual context initially improves the direct AV-in-context baseline, performance sharply degrades past 32 frames, plummeting to an average score of 22.0 at 1024 frames due to memory exhaustion. This catastrophic failure highlights the limitations of standard omnimodal models when processing excessively long joint token sequences, which typically lead to substantial memory overhead and inference bottlenecks. Token compression baselines (OmniZip, OMAC, OMARC) extend this context ceiling by compressing the input sequence and establishing a robust memory distillation. However, they ultimately collapse because maintaining a viable context window across hundreds of frames demands aggressive pruning, which discards sparse but critical cross-modal anchors, decouples subtle acoustic cues from their corresponding visual events, and leads to modality collapse. \texttt{Omni2LoRA} completely avoids the active token-space bottleneck by internalizing the temporal context directly into parametric memory. Rather than collapsing under high frame counts, our approach uniquely capitalizes on the expanded cross-modal evidence, scaling gracefully and monotonically to reach a peak accuracy of 46.2 at 1024 frames.

\paragraph{Inference Efficiency and Amortization.} VidCapBench provides a rigorous setting to evaluate inference efficiency by associating multiple sequential queries with a single video context. Figure \ref{fig:amortization}(a) shows that \texttt{Omni2LoRA} massively reduces the average Time to First Token (TTFT). On the 7B backbone, it averages 0.49s per question, outperforming both the full-context baseline (6.03s) and the strongest token-compression baseline, OMARC (3.45s).This performance gap stems from how different frameworks handle iterative querying. Traditional pruning methods (e.g., OMAC and OmniZip) retain visual tokens in the context, forcing repeated encoding of multimodal inputs for every subsequent query. Conversely, \texttt{Omni2LoRA} incurs a one-time setup cost to internalize the video into a reusable parametric adapter, driving downstream multimodal token loads to zero. Figure \ref{fig:amortization}(b) tracks this dynamic over sequential questions: while OMARC suffers continuously high latency (3.6s–4.2s for 7B), \texttt{Omni2LoRA}'s amortized TTFT drops precipitously. After just five queries, effective latency falls to 0.82s (7B) and 0.72s (3B), ultimately plateauing near 0.43s. Translating video memory into a fixed-budget parametric adapter eliminates the repeated per-query bottleneck.

\section{Conclusion}
\label{sec:conclusion}

We introduce \texttt{Omni2LoRA}, a two-stage framework that bypasses the token bottleneck by compressing audio--visual recordings into parametric memory. A Perceiver hypernetwork first internalizes each recording into LoRA weights, while a GRPO-trained allocation policy compresses them into a fixed, sub-linear rank budget using a coherence-aware advantage that preserves cross-modal dependencies. \texttt{Omni2LoRA} at a 30\% rank budget outperforms direct full-context inference and state-of-the-art token compression baselines, remains robust under compression ratios up to 75\%, and scales gracefully to long recordings where token-pruning methods degrade sharply. By eliminating multimodal tokens from the active context at inference time, our method reduces per-query TTFT by an order of magnitude and amortizes to sub-second latency across repeated queries. Future work will explore more scalable adapter generation and allocation strategies for streaming long-form multimodal content.

\bibliography{aaai2027}
\newpage
\appendix
\section{Supplementary Materials}
\label{sec:appendix}

\section{A\quad Limitations and Ethics Statement}
\label{app:checklist}

\subsection{Limitations}
\label{app:limitations}
\begin{itemize}
\item \textbf{One adapter per recording.} Memory is recording-specific;
answering across multiple recordings requires either re-internalization or an
adapter-composition mechanism, which we do not study.
\item \textbf{Non-streaming.} Internalization assumes the full recording is
available before querying, so the method as presented does not support live
streams.
\item \textbf{Setup cost dominates at $n=1$.} The efficiency argument depends on
repeated querying; for a single query, $T_{\text{setup}}$ is not amortized.
\item \textbf{Two-stage dependence.} Stage~2 can only select among directions
Stage~1 produced; information absent from the candidate bank is unrecoverable.
\item \textbf{Rule-based reward.} $R_{\text{acc}}$ and $R_{\text{fmt}}$ are
cheap but coarse, and may under-credit correct free-form answers.
\item \textbf{Scope of evidence.} Results cover three backbones, five
benchmarks and English-language content.
\end{itemize}
% =====================================================================
\subsection{Ethics and Broader Impact}
\label{app:ethics}
Reducing per-query latency and eliminating repeated multimodal encoding lowers
the energy cost of serving audio--visual assistants and makes long-recording
understanding feasible on smaller deployments. A recording-specific adapter is a compressed encoding of that recording's content, and should be treated as carrying the same privacy sensitivity and the same retention and deletion policy as the source media as caching adapters is not equivalent to discarding the video. We use only publicly released research datasets, under their stated licenses and intended research use, and introduce no new human-subjects data or annotation.
% =====================================================================
\section{B\quad Method Details and Pseudocode}
\label{app:method}
This section addresses Checklist item~1.1. We restate the notation, give
complete pseudocode for both training stages and for inference, specify the
architecture of every trainable component, derive the allocation likelihood
and its gradient, and define the reward terms exactly as implemented.
\subsection{B.1\quad Notation}
\label{app:notation}
\begin{table}[h]
\centering
\small
\setlength{\tabcolsep}{4pt}
\begin{tabular}{l p{0.62\columnwidth}}
\toprule
\textbf{Symbol} & \textbf{Meaning} \\
\midrule
$v$, $v_t$ & Synchronized audio--visual recording; its $t$-th temporal chunk \\
$T$ & Number of non-overlapping temporal chunks \\
$i$ & Internalization instruction (fixed template, App.~H.1) \\
$q$, $y$ & Downstream text query; target response \\
$E$, $F$ & Frozen OLM encoder; frozen OLM answer model \\
$C_t$ & Layer-wise hidden states of chunk $t$, $\mathbb{R}^{L\times S\times D}$ \\
$L, M$ & \# transformer layers; \# target linear modules per layer \\
$S$, $D$ & Fused audio--visual sequence length; hidden dimension \\
$H_\phi$ & Perceiver hypernetwork (trained in Stage~1, frozen in Stage~2) \\
$R_{\max}$ & Candidate rank directions generated per $(\ell,m,t)$ triple \\
$\mathcal{K}$ & Full slot bank, $|\mathcal{K}| = T\cdot L\cdot M\cdot R_{\max}$ \\
$\mathcal{K}_{\text{sel}}$ & Selected allocation, $|\mathcal{K}_{\text{sel}}| = R_{\text{total}}$ \\
$g_\psi$ & Scoring network (trained in Stage~2) \\
$z_{\ell,m,t,r}$ & Scalar logit assigned to slot $(\ell,m,t,r)$ \\
$\rho$ & Retained ratio $R_{\text{total}}/|\mathcal{K}|$; compression ratio $= 1-\rho$ \\
$s$ & Fixed LoRA scaling factor \\
$G$ & Group size (rollouts per query) for GRPO \\
$\Omega(q)$ & Audio--Visual Dependence Score of query $q$ \\
$\eta,\lambda,\beta$ & Coherence penalty strength; shaping weight; KL coefficient \\
$\tau$ & Numerical-stability constant \\
\bottomrule
\end{tabular}
\caption{\small{Notation used throughout the paper and appendix.}}
\label{tab:notation}
\end{table}
\subsection{B.2\quad Stage 1: Full-Rank Hypernetwork Training}
\label{app:alg_stage1}
\begin{algorithm}[h]
\caption{Stage 1 --- Full-Rank Hypernetwork Training}
\label{alg:stage1}
\begin{algorithmic}[1]
\REQUIRE Caption corpus $\mathcal{D}_1=\{(v,i,p,y)\}$; frozen encoder $E$ and answer model $F$; hypernetwork $H_\phi$; chunk length $\Delta$; rank cap $R_{\max}$; scaling $s$
\ENSURE Converged, frozen hypernetwork $H_\phi$
\STATE Initialize $\phi$ \COMMENT{App.~B.5}
\FOR{each optimizer step}
\STATE Sample minibatch $\{(v,i,p,y)\}\subset\mathcal{D}_1$
\FORALL{$v$ in minibatch}
\STATE Partition $v$ into $T=\lceil \mathrm{dur}(v)/\Delta \rceil$ chunks $\{v_1,\dots,v_T\}$
\FOR{$t=1$ \TO $T$}
\STATE $C_t \leftarrow E(v_t, i)$ \COMMENT{no gradient through $E$}
\STATE $\{A_{\ell,m,t,r},B_{\ell,m,t,r}\}_{r=1}^{R_{\max}} \leftarrow H_\phi(C_t)$
\ENDFOR
\STATE Compose the full adapter $\theta(v)$ over all $T\cdot L\cdot M\cdot R_{\max}$ rank-one components and inject via Eq.~\eqref{eq:lora-inject}
\STATE Compute teacher-forced loss $\mathcal{L}(\phi)$ (Eq.~\eqref{eq:stage1-loss}) with $F$ frozen
\ENDFOR
\STATE $\phi \leftarrow \mathrm{AdamW}\big(\phi, \nabla_\phi \mathcal{L}\big)$
\ENDFOR
\STATE Freeze $\phi$; $H_\phi$ is now a deterministic map $v \mapsto \mathcal{K}$
\STATE \textbf{return} $H_\phi$
\end{algorithmic}
\end{algorithm}
\noindent Gradients flow only into $\phi$; $E$ and $F$ are frozen and run in
inference mode (no dropout, no parameter updates) throughout. Teacher captions
are generated \emph{offline} once and cached, so no teacher model is loaded
during Stage~1 optimization.
\subsection{B.3\quad Stage 2: O2L-GRPO}
\label{app:alg_stage2}
\begin{algorithm}[h]
\caption{Stage 2 --- Coherence-Aware Rank Allocation (O2L-GRPO)}
\label{alg:stage2}
\begin{algorithmic}[1]
\REQUIRE Query corpus $\mathcal{D}_2=\{(v,q,y)\}$; frozen $E$, $F$, $H_\phi$; scoring net $g_\psi$; budget $R_{\text{total}}$; group size $G$; coefficients $\eta,\lambda,\beta$; clip range $\epsilon$
\ENSURE Converged, frozen scoring network $g_\psi$
\STATE \textbf{// Phase 0: cache frozen unimodal references (once)}
\FORALL{$(v,q,y)\in\mathcal{D}_2$}
\STATE $R^{\text{AV}} \leftarrow \mathrm{Reward}\big(F(q \mid \text{full AV context}), y\big)$
\STATE $R^{\text{V}} \leftarrow \mathrm{Reward}\big(F(q \mid \text{audio masked}), y\big)$
\STATE $R^{\text{A}} \leftarrow \mathrm{Reward}\big(F(q \mid \text{frames masked}), y\big)$
\STATE $\Omega(q) \leftarrow \mathrm{ReLU}\big(R^{\text{AV}} - \max(R^{\text{V}},R^{\text{A}})\big)$
\STATE Cache $(R^{\text{AV}},R^{\text{V}},R^{\text{A}},\Omega(q))$
\ENDFOR
\STATE $\psi_{\text{ref}} \leftarrow \psi$ \COMMENT{reference policy for KL}
\STATE \textbf{// Phase 1: policy optimization}
\FOR{each optimizer step}
\STATE Sample $(v,q,y)\sim\mathcal{D}_2$
\STATE $\{C_t\}_{t=1}^{T}\leftarrow E(v,i)$; $\mathcal{K}\leftarrow H_\phi(\{C_t\})$ \COMMENT{cacheable per $v$}
\STATE $z_{j} \leftarrow g_\psi(\bar{C}_{t(j)},\ell(j),m(j),r(j))$ for all $j\in\mathcal{K}$
\FOR{$i=1$ \TO $G$}
\STATE $\mathcal{K}^{(i)}_{\text{sel}} \leftarrow \textsc{SampleWoR}(z, R_{\text{total}})$ \COMMENT{Eq.~\eqref{eq:pl-step}}
\STATE Build adapter from $\mathcal{K}^{(i)}_{\text{sel}}$; $\hat{y}_i \leftarrow F(q\mid \text{zero AV tokens})$
\STATE $R^{\text{alloc}}_i \leftarrow \tfrac12 R_{\text{acc}}(\hat y_i,y) + \tfrac12 R_{\text{fmt}}(\hat y_i)$
\ENDFOR
\STATE $\hat{A}_i \leftarrow \big(R^{\text{alloc}}_i - \mu_R\big)/(\sigma_R+\tau)$, with $\mu_R,\sigma_R$ the group mean/std
\STATE $\Delta_i \leftarrow \mathrm{ReLU}(R^{\text{AV}}-R^{\text{alloc}}_i)/(|R^{\text{AV}}|+\tau)$
\STATE $\tilde{\Delta}_i \leftarrow \Delta_i \cdot (1+\eta\cdot\Omega(q))$
\STATE $w_i \leftarrow \mathrm{ReLU}(\hat{A}_i)\cdot\tilde{\Delta}_i$; \quad $\tilde{A}_i \leftarrow \hat{A}_i - \lambda w_i$
\STATE $\psi \leftarrow \mathrm{AdamW}\big(\psi, \nabla_\psi \mathcal{L}_{\text{O2L-GRPO}}\big)$ (Eq.~\eqref{eq:final-obj})
\ENDFOR
\STATE \textbf{return} frozen $g_\psi$
\end{algorithmic}
\end{algorithm}
\noindent Because $F$, $E$ and $H_\phi$ are all frozen, both the unimodal
reference rewards and the slot bank $\mathcal{K}$ for a given recording are
deterministic and are computed once and cached; the only stochasticity in the
loop comes from allocation sampling (and, if enabled, decoding).
\subsection{B.4\quad Inference}
\label{app:alg_infer}
\begin{algorithm}[h]
\caption{Inference with \texttt{Omni2LoRA}}
\label{alg:inference}
\begin{algorithmic}[1]
\REQUIRE Recording $v$; query stream $q_1,\dots,q_N$; frozen $E,H_\phi,g_\psi,F$; budget $R_{\text{total}}$
\STATE \textbf{// One-time internalization}
\STATE $\{C_t\}\leftarrow E(v,i)$; $\mathcal{K}\leftarrow H_\phi(\{C_t\})$
\STATE $z_j \leftarrow g_\psi(\cdot)$ for all $j\in\mathcal{K}$
\STATE $\mathcal{K}_{\text{sel}} \leftarrow \textsc{TopK}(z, R_{\text{total}})$ \COMMENT{greedy, no sampling}
\STATE $\theta_{\text{mem}}(v) \leftarrow$ adapter assembled from $\mathcal{K}_{\text{sel}}$; cache to disk
\STATE \textbf{// Per-query answering (zero multimodal tokens in context)}
\FOR{$n=1$ \TO $N$}
\STATE $\hat{y}_n \leftarrow F\big(q_n \mid \theta_{\text{mem}}(v)\big)$
\ENDFOR
\STATE \textbf{return} $\{\hat{y}_n\}$
\end{algorithmic}
\end{algorithm}

This section details the inference procedure outlined in Algorithm~\ref{alg:inference}. The inference sequence is systematically partitioned into a one-time internalization phase and a sequential querying phase. During the initial internalization, the frozen multi-modal encoder and the Perceiver hypernetwork evaluate the input media to generate a deterministic pool of candidate low-rank adapter slots. Subsequently, the scoring network evaluates these parameters to deterministically isolate the top-$R_{\text{total}}$ slots, assembling a highly compressed, video-specific memory adapter which is persistently cached. During the secondary querying phase, any downstream text prompt is processed by the frozen answer model augmented solely by this cached memory adapter. This execution necessitates zero multi-modal tokens within the context window, effectively decoupling the inference latency from the temporal length of the original recording.

\subsection{B.5\quad Hypernetwork Architecture ($H_\phi$)}
\label{app:hypernet}
$H_\phi$ is a hierarchical Perceiver-style resampler that cross-attends a set
of learned latent queries into the frozen layer-wise states $C_t$ and projects
the resulting latents into rank-one LoRA factors.

\subsection{B.6\quad Scoring Network Architecture ($g_\psi$)}
\label{app:scorer}
$g_\psi$ consumes (i) the pooled chunk summary $\bar{C}_t=\mathrm{mean}(C_t)$
and (ii) a positional embedding of the slot index $(\ell,m,r)$, and emits a
single logit. It never reads the continuous factors $A,B$, which keeps the
Stage~2 action space purely combinatorial.
\subsection{B.7\quad Slot Bank and Budget Accounting}
\label{app:budget}
Every slot is a rank-one update, so the stored adapter cost is
\begin{equation}
\small
\mathrm{Params}(\mathcal{K}_{\text{sel}}) = \sum_{(\ell,m,t,r)\in\mathcal{K}_{\text{sel}}} \big(d^{(\ell,m)}_{\text{in}} + d^{(\ell,m)}_{\text{out}}\big),
\end{equation}
and the full candidate bank contains $|\mathcal{K}| = T\cdot L\cdot M\cdot R_{\max}$ slots.
The \emph{retained ratio} reported in all tables is
\begin{equation}
\small
\rho = \frac{R_{\text{total}}}{|\mathcal{K}|}, \qquad \text{compression ratio} = 1-\rho .
\end{equation}
In Tables~\ref{tab:main_comparison} and \ref{tab:main_results} and
Fig.~\ref{fig:comp_ratio}, $\rho$ is held fixed at the stated value with a
32-frame budget.
\subsection{B.8\quad Allocation Likelihood and Gradient}
\label{app:derivation}
For completeness we give the self-contained derivation of the quantity
differentiated in Eq.~\eqref{eq:final-obj}. This is an implementation detail,
not a theoretical claim (see App.~E).
Sampling $R_{\text{total}}$ distinct slots sequentially without replacement,
with the step-wise conditional of Eq.~\eqref{eq:pl-step}, induces the joint
\begin{equation}
\small
\pi_\psi(\mathcal{K}_{\text{sel}}\mid v) = \prod_{k=1}^{R_{\text{total}}} \frac{\exp(z_{o_k})}{\sum_{j\in\mathcal{K}\setminus o_{<k}}\exp(z_j)} .
\end{equation}
Taking logarithms turns the product into a sum of per-step terms,
\begin{equation}
\small
\begin{split}
\log \pi_\psi(\mathcal{K}_{\text{sel}}\mid v) = \sum_{k=1}^{R_{\text{total}}}\Big[ z_{o_k} - \log\sum_{j\in\mathcal{K}\setminus o_{<k}}\exp(z_j) \Big],
\end{split}
\end{equation}
which is exactly the log-likelihood of an autoregressive sequence model over
$R_{\text{total}}$ ``tokens'' drawn from a shrinking vocabulary. Differentiating
the $k$-th term with respect to a logit $z_u$ gives
\begin{equation}
\small
\frac{\partial}{\partial z_u}\log\pi_\psi = \sum_{k=1}^{R_{\text{total}}} \Big[ \mathbf{1}[u=o_k] - \mathbf{1}[u\notin o_{<k}]\pi^{(k)}_u \Big],
\end{equation}
where $\pi^{(k)}_u$ is the step-$k$ softmax probability of slot $u$ over the
surviving candidate set. The chain rule then propagates into $\psi$ through
$z=g_\psi(\cdot)$. Two consequences matter in practice:
\begin{enumerate}
\item The factorization is a valid probability over ordered selections, so the
importance ratio $\rho_i(\psi)=\pi_\psi(\mathcal{K}^{(i)}_{\text{sel}}\mid v)/\pi_{\psi_{\text{old}}}(\mathcal{K}^{(i)}_{\text{sel}}\mid v)$ is well defined and PPO-style clipping applies unchanged to this combinatorial action space.
\item The normalizer is computed with a running mask over already-selected
slots; we accumulate log-probabilities in log-space and mask with $-\infty$
rather than renormalizing explicitly, for numerical stability.
\end{enumerate}
The KL term is estimated as against the fixed reference policy
$\pi_{\psi_{\text{ref}}}$.
\subsection{B.9\quad Reward Specification}
\label{app:reward}
All rewards are rule-based; no auxiliary judge model is invoked during RL
sampling. With $R=\tfrac12 R_{\text{acc}} + \tfrac12 R_{\text{fmt}}$:
\begin{itemize}
\item $R_{\text{acc}}\in\{0,1\}$: for multiple-choice items, $1$ iff the parsed
option letter matches the gold letter after normalization (case folding,
stripping of punctuation and leading articles).
\item $R_{\text{fmt}}\in\{0,1\}$: $1$ iff the completion matches the required
output template (App.~H.2), i.e.\ contains exactly one well-formed answer
delimiter pair and no trailing content.
\item Ties, empty generations, and parse failures receive $R_{\text{acc}}=0$
and $R_{\text{fmt}}=0$.
\end{itemize}
The same reward function is used for the three cached unimodal references
($R^{\text{AV}},R^{\text{V}},R^{\text{A}}$) and for the student rollouts
$R^{\text{alloc}}_i$, so that $\Delta_i$ compares like with like.

% =====================================================================
\section{C\quad Background for Less-Familiar Readers}
\label{app:background}
This section addresses Checklist item~1.3. Readers new to any component of the
pipeline may find the following entry points useful.
\begin{itemize}
\item \textbf{Low-Rank Adaptation.} Adapting a frozen weight matrix by an
additive low-rank product, rather than by updating the matrix itself, is
introduced in \cite{hu2022lora}; this is the parameterization we generate
rather than learn by gradient descent.
\item \textbf{Hypernetworks.} Networks that emit the weights of another network
in a single forward pass are introduced in \cite{ha2017hypernetworks}; the
context-to-adapter instantiations most relevant here are
\cite{charakorn2026doc,suri2026frames2lora}.
\item \textbf{Perceiver resampling.} Cross-attending a small set of learned
latents into a long input sequence, giving cost linear rather than quadratic in
input length, is described in \cite{jaegle2021perceiver}. This is what allows
$H_\phi$ to consume long fused audio--visual states.
\item \textbf{Policy-gradient RL with clipped ratios.} The clipped surrogate
objective is due to \cite{schulman2017proximal}; the group-relative,
critic-free variant we adapt is \cite{shao2024deepseekmath}.
\item \textbf{Omnimodal language models and token compression.} For the
backbone paradigm see \cite{cheng2024videollama,xu2025qwen3,liu2026javisgpt};
for the token-reduction literature we compare against, see
\cite{bolya2022token,chen2024image,tao2025omnizip,wu2026marc}.
\item \textbf{Audio--visual correspondence.} Why isolated per-modality
compression is lossy for joint reasoning is discussed in
\cite{tian2018audio,li2022learning}.
\end{itemize}
% =====================================================================
\section{D\quad Scope of Claims}
\label{app:claims}
This section addresses Checklist item~1.2 by separating what we measure from
what we interpret.
\paragraph{Empirically established (measured, reported with variability).}
(i) At a 30\% retained ratio, \texttt{Omni2LoRA} attains higher average accuracy
than OmniZip, OMAC, O-MARC and the full-token baseline on all three backbones
(Table~\ref{tab:main_comparison}). (ii) Accuracy degrades more slowly than all
token-pruning baselines as the compression ratio increases to 75\%
(Fig.~\ref{fig:comp_ratio}). (iii) Accuracy is non-decreasing in frame count up
to 1024 frames, where in-context baselines degrade
(Fig.~\ref{fig:frame_ablation}). (iv) Per-query TTFT and its amortization
profile (Fig.~\ref{fig:amortization}). (v) The ablation ordering
O2L-GRPO $>$ Norm-scored $>$ Uniform at matched budget
(Table~\ref{tab:main_results}).
% =====================================================================
\section{E\quad Theoretical Contributions}
\label{app:theory}
This section addresses Checklist items~2.1--2.8. The paper makes \textbf{no
theoretical contributions}: we state no theorems, prove no novel claims, and
derive no bounds. Consequently items 2.2--2.8 are \textbf{NA}.
The algebra in App.~B.8 is included for reproducibility --- it specifies exactly
which log-likelihood is differentiated --- and not as a novel result; it is a
direct consequence of the sampling scheme defined in Eq.~\eqref{eq:pl-step}.
All claims in the paper are empirical and are supported by the experiments in
App.~G.
% =====================================================================
\section{F\quad Datasets}
\label{app:data}
This section addresses Checklist items~3.1--3.7.
\subsection{F.1\quad Training Corpora and Motivation}
\label{app:train_data}
\paragraph{Stage 1 --- VALOR-1M \cite{liu2024valor}.} Chosen because it provides
recordings with \emph{jointly} annotated audio and visual content at scale,
which is the supervision signal required to teach $H_\phi$ to compress both
streams into a single adapter; visual-only caption corpora would bias the
hypernetwork toward exactly the visual dominance that Stage~2 must later
correct.
\paragraph{Stage 2 --- FineVideo \cite{farre2024finevideo}.} Chosen for its
long, naturalistic recordings with rich co-occurring speech, ambient sound and
scene structure, which yields a non-degenerate distribution of the dependence
score $\Omega(q)$; corpora where audio is redundant with vision would leave the
coherence penalty inactive. Queries and video references are additionally drawn
from downstream \emph{training} splits to align the allocation policy with the
evaluation query distribution.
\subsection{F.2\quad Evaluation Benchmarks}
\label{app:eval_data}
All evaluation datasets are drawn from existing literature, are cited in-line,
and are publicly available (Checklist 3.5, 3.6).

\begin{table*}[t]
\centering
\small
\setlength{\tabcolsep}{5pt}
\begin{tabular}{l c c c l}
\toprule
\textbf{Benchmark} & \textbf{\#Videos} & \textbf{\#Queries} & \textbf{Metric} & \textbf{License} \\
\midrule
UGC-AVQA \cite{wu2026marc}                  & 206      & 1{,}648   & Acc.\ (\%)  & Research-only (links + annotations) \\
WorldSense \cite{hong2025worldsense}        & 1{,}662  & 3{,}172   & Acc.\ (\%)  & Not specified \\
OmniVideoBench \cite{li2025omnivideobench}  & 628      & 1{,}000   & Avg.\ score & CC BY-NC-ND 4.0 (gated) \\
DailyOmni \cite{zhou2025daily}              & 684      & 1{,}197   & Acc.\ (\%)  & CC BY-NC-SA 4.0 \\
VidCapBench \cite{chen2025vidcapbench}      & 643      & 9{,}494   & TTFT (s)    & Custom, academic-only \\
\midrule
VALOR-1M \cite{liu2024valor} (train)        & $\sim$1M & ---       & ---         & MIT (code/annot.); videos via AudioSet \\
FineVideo \cite{farre2024finevideo} (train) & 43{,}751 & $\sim$219K & --- & CC-BY (per-video) \\
\bottomrule
\end{tabular}
\caption{\small{Datasets used, with the split evaluated and the license under
which each is distributed. UGC-AVQA figures are for the difficulty-filtered
benchmark split. WorldSense, OmniVideoBench, DailyOmni, and
VidCapBench are each distributed as a single undivided release; we evaluate the
full release in every case. VidCapBench counts refer to the
automatically-assessable (AE) subset; the 1{,}150-pair human-assessable (HE)
subset is not used. Snapshots: WorldSense HuggingFace commit
\texttt{49df5aa}, DailyOmni commit \texttt{bf5a6ee}; OmniVideoBench and
FineVideo are gated repositories accessed under their respective terms of use.}}
\label{tab:datasets}
\end{table*}

\noindent UGC-AVQA is the discriminative test: its four
categories (Audio--Visual Event Progression, Scene/Temporal Transition,
Cross-Scene Audio--Visual Alignment, Fine-Grained Audio--Visual Contrast)
strictly require joint evidence, so it isolates coherence preservation rather
than general video competence. Its construction reinforces this: a candidate is
retained only if a strong reference model fails on the same question once the
audio track is removed, which filters out items solvable from visual priors
alone. WorldSense and DailyOmni test that gains generalize to broad and to
everyday-scenario omnimodal understanding. OmniVideoBench stresses long-horizon
reasoning, where a parametric memory should help most. VidCapBench is used
solely for efficiency, because it pairs many sequential queries with a single
recording, which is precisely the setting in which one-time internalization
amortizes.

\paragraph{Availability.}
All datasets are obtainable from the following locations:
\begin{description}
  \raggedright
  \item[UGC-AVQA:] \url{https://github.com/WPR001/UGC_VideoCaptioner}
  \item[WorldSense:] \url{https://huggingface.co/datasets/honglyhly/WorldSense}
  \item[OmniVideoBench:] \url{https://huggingface.co/datasets/NJU-LINK/OmniVideoBench}
  \item[DailyOmni:] \url{https://huggingface.co/datasets/liarliar/Daily-Omni}
  \item[VidCapBench:] \url{https://huggingface.co/datasets/VidCapBench/VidCapBench}
  \item[VALOR-1M:] \url{https://github.com/CASIA-IVA-Lab/VALOR}
  \item[FineVideo:] \url{https://huggingface.co/datasets/HuggingFaceFV/finevideo}
\end{description}

\section{G\quad Computational Experiments}
\label{app:experiments}

\subsection{G.1\quad Evaluation Metrics}
\label{app:metrics}
\paragraph{Overall accuracy (\%).} Fraction of instances scored correct by the
benchmark's own protocol, used for DailyOmni, UGC-AVQA and WorldSense. Chosen
because these are multiple-choice benchmarks whose official leaderboards report
accuracy, keeping our numbers directly comparable to published baselines.
\paragraph{Average score.} OmniVideoBench's official aggregate, retained rather
than converted to accuracy so that our numbers remain comparable to that
benchmark's reported results.
\paragraph{Per-category accuracy (UGC-AVQA).} AVEP, AVST, CSAVA and FGAVC are
reported separately because our central claim is about \emph{cross-modal}
preservation; an aggregate could hide a gain on visually-solvable items paired
with a loss on strictly joint ones.
\paragraph{Retained ratio and compression ratio.} Defined in App.~B.7. We match
the retained ratio, not the wall-clock cost, across methods so that comparisons
isolate the effect of \emph{what} is kept rather than \emph{how much} compute is
spent.
\paragraph{Time to First Token (TTFT).} Wall-clock seconds from query submission
to emission of the first output token, measured in ms.
Single-question TTFT in Fig.~\ref{fig:amortization}(a) \emph{includes} the
one-time internalization cost, so the comparison is not favorable to us by
construction.
\paragraph{Amortized TTFT.} For $n$ sequential queries against one recording,
\begin{equation}
\small
\mathrm{TTFT}_{\text{amort}}(n) = \frac{T_{\text{setup}} + \sum_{j=1}^{n} t_j}{n},
\end{equation}
where $T_{\text{setup}}$ is the one-time encode-plus-allocate cost (zero for
in-context baselines) and $t_j$ is the per-query TTFT. This metric is chosen
because it is the quantity a deployment actually pays under repeated querying,
and it penalizes our method at $n=1$.

\end{document}